\documentclass[runningheads]{llncs}
\usepackage{cite}
\usepackage{amsmath,amssymb,amsfonts}
\usepackage{algorithmic}
\usepackage{textcomp}
\usepackage{xcolor}
\usepackage{enumitem}
\usepackage{lipsum, graphics, subcaption}

\usepackage[utf8]{inputenc}  
\usepackage{wrapfig}
\usepackage{multirow}
\usepackage{hyperref}  
\newcommand{\mat}[1]{\mathbf{#1}} 

\begin{document}
\title{Linear Self-Attention Approximation via Trainable Feedforward Kernel}
\titlerunning{Linear Self-Attention Approximation}
%
\author{Uladzislau Yorsh \and
Alexander Kovalenko\orcidID{0000-0002-7194-1874}}
\authorrunning{Yorsh et al.}
%
\institute{Faculty of Information Technology, Czech Technical University in Prague\\
Prague, Czech Republic\\
\email{\{yorshula, kovalale\}@fit.cvut.cz}}
\maketitle              
%

%
%
%

Restrictive limitation of Transformers\cite{vaswani2017attention} due to the quadratic complexity of self-attention mechanism motivated a new research field of \textit{efficient Transformers}\cite{tay2020efficient}, which approximate the original architecture with asymptotically faster models.


Despite the fact that Transformers are pervasive, unbiased and able to virtually handle arbitrarily long dependencies, the quadratic space and time complexity limit Transformer applications on long sequences. In this connection, various findings on approximating the attention with asymptotically faster modules have been made in order to tackle longer sequences. However, given the absence of a unified and systematic benchmark, overall evaluating remained uncertain until Tay et al. \cite{tay2020long} published the benchmark for efficient Transformer models called "Long Range Arena", that consists of task of various data types.


In pursue of the faster computation, Efficient Transformers demonstrate an impressive variety of approaches---models attaining sub-quadratic attention complexity can utilize a notion of sparsity \cite{roy2020efficient, child2019sparse, beltagy2020longformer} or a low-rank approximation of inputs \cite{linformer, nystromformer} to reduce an amount of attended keys; another ways to reduce complexity include locality-sensitive hashing\cite{kitaev2020reformer}, key pooling\cite{poolingformer}, additional memory to store information in compacted form\cite{transformer_xl, rae2019compressive} or hybridization with another architectures, such as CNNs\cite{gulati2020conformer, bello2020attention}.


Often based on strong mathematical basis, kernelized approaches allow to approximate an attention with linear complexity while retaining high accuracy. The work by Katharopoulos et al. \cite{katharopoulos2020transformers} describes an approximation consisting of computing an attention by a dot product of projected queries and keys. Consequently, the work by Choromanski et al.k\cite{choromanski2021rethinking} demonstrated that such an approximation can be arbitrarily precise and mathematically robust, while the work by Chowdhury et al. \cite{chowdhury2021learning} reported that the projection can be learned. Therefore, in the present paper we aim to expand the idea of trainable kernel methods to approximate self-attention mechanism of the Transformer architecture.






\textbf{Our contribution:} given that feedforward neural network with at least one hidden layer \cite{cybenko1989approximation}, arbitrary non-linearity, and arbitrary number of neurons is able to approximate any well-behaved function to any accuracy that gives feedforward neural network the potential of being universal approximator \cite{hornik1991approximation}. Therefore, we propose that trainable kernel function $\phi(\cdot)$ can approximate traditional softmax attention efficiently. Therefore, we study the possibility of using the feedforward neural network to represent $\phi(\cdot)$. We experiment with the architecture of $\phi(\cdot)$ and test its performance on the three Long Range Arena tasks---text classification, document matching and ListOps, following the instruction on limitation \cite{tay2020long} the number of trainable parameters in the model to provide comparable metrics. 

\textbf{Kernelized Model.} Kernelized models are based on the following factorization of an attention operation:

\[\text{Att}(q_i, K, V) = \sum_{j=1}^{L} \frac{\kappa(q_i, k_j)}{\sum_{j'=1}^{L} \kappa(q_i, k_{j'})} v_j \approx
\frac{\phi(q_i)^T \sum_{j=1}^{L} \phi(k_j) v_j^T }{\phi(q_i)^T \sum_{j=1}^{L} \phi(k_j)}\]
where $q_i$ is a query token, $K$ and $V$ are key and value matrices, $\kappa(q, k)$ is a kernel function to model ($\exp{(q^Tk)}$ for a base Transformer) and $\phi(\cdot)$ is a projection function we approximate. The $\phi(\cdot)$ is required to be positive to maintain numeric stability, and can vary from simple functions like ELU + 1 to stochastic softmax kernel approximations. In our work, instead of approximation strategies with strong priors we employ a general function such as feedforward NN.

\textbf{Feedforward Kernel.} We start with a single-layer FFN, defined as:

\[\phi(\mat X) = Softplus(\mat X \mat W)\]
\\
where $W \in R^{n \times n}$ is the layer weight matrix. Surprisingly, this model already shows the notable performance gain over the Performer and comes close to the leader of the original LRA paper. Following the \cite{jia2019orthogonal}, we can boost performance by forcing the orthogonality via orthogonal initialization and regularization.

We also tried to stack more layers, but observed no performance gain -- with or without normalization layers in between. We tried the GELU and logistic sigmoid non-linearities.

\textbf{GLU Kernel.} Gated Linear Units are defined as:

\[GLU(\mat X) = \mat X \mat W_f \odot \sigma(\mat X \mat W_g)\]
\\
where $\sigma(\cdot)$ is a logistic sigmoid and $W_f$, $W_g$ are weight matrices. This layer provides and element-wise nonlinearity and may represent more complex functions, but requires a doubled parametrization compared to a linear one.

For the purposes of our model, we need to modify the last GLU to force the positive output:

\[GLU_{output}(\mat X) = Softplus(\mat X \mat W_f) \odot \sigma(\mat X \mat W_g)\]
\\
We also force the orthogonality of $W_f$ in these units in the same way as in the previous subsection. We refer this regularized units as O(rthogonal)GLU.


To mitigate the parametrization growth we apply transforms head-wise, and suggest that gating does not require that amount of information as the input transform. Thus, we can approximate the $W_g$ with, say, two low-rank matrices of sizes $n \times r$ resp. $r \times n$ where $r < \frac{n}{2}$ is the $W_g$ approximation rank. We refer this unit as A(pproximated)OGLU.


\textbf{Gating.} Compared to the orthogonal single-layer FFN, a single-layer OGLU model converges even faster and shows significantly less score variance between runs. These units can also be sequentially stacked with a benefit, up to some extent. On the other hand, the doubled parametrization will not allow stacking more than two units without going beyond the 10\% of additional parameters.

By approximating the gating weight matrix, we are able to stack more units -- but according to the Table \ref{tab:results}, this brings no advantage with the higher computational costs. We used the matrices of rank $r = n/4$ to approximate the gate, reducing the layer parametrization by 25\%.

\textbf{Experiments.} Following the recommendations from \cite{tay2020long}, we replicate the learning schedule and all the hyperparameters that relate to our model, while keeping additional parametrization below 10\% Due to the limitation in computational power, we restrict ourselves only to the three LRA tasks---BPE text classification, BPE text matching and ListOps, with input lengths $4K/4K/2K$ respectively. To provide comparable and reproducible and results, we used the gradient accumulation in order to simulate larger batch sizes. Each model was trained five times to observe model behavior and to avoid so-called \textit{black swans}---random seeds that give radically different results \cite{picard2021torchmanualseed3407}. Mean and best results are reported in Table \ref{tab:results}



\begin{center}
    \begin{table}
    \centering
    \begin{tabular}{l | c | c | c | c }
        \hline
        & & & \\
        Model & Complexity & Classif. & Matching & ListOps\\
        & & & \\
        \hline\hline
        Transformer & $\mathcal{O}(L^2)$ & 64.27 & 57.46 & 36.37\\
        \hline
         Linear kernel\textsuperscript{\textbf{\dag}} & $\mathcal{O}(CL)$ & 65.77 & 73.51 & 18.54\\
         $1 \times$ GLU & $\mathcal{O}(CL)$ & 65.82 & 72.17 & 18.67 \\
         $2 \times$ GLU & $\mathcal{O}(CL)$ & 65.99 & 73.36 & 18.42 \\
         $3 \times$ GLU & $\mathcal{O}(CL)$ & 65.87 & 72.60 & 18.68 \\
         & & & & \\
         Orth. linear kernel & $\mathcal{O}(CL)$ & 65.86 & 72.63 & 18.19\\
         $1 \times$ OGLU & $\mathcal{O}(CL)$ & 65.95 & 72.50 & 18.45\\
         $2 \times$ OGLU & $\mathcal{O}(CL)$ & 66.02 & 72.96 & 18.32\\
         $3 \times$ AOGLU& $\mathcal{O}(CL)$ & 66.06 & 72.57 & 18.45\\
        \hline
    \end{tabular}
    \caption{Results of our models on the chosen LRA tasks, mean results for five runs. We denote by \textbf{\dag} models that show significant variance in results.}
    \label{tab:results}
    \end{table}
\end{center}

\section*{Acknowledgment}
This research is supported by the Czech Ministry of Education, Youth and Sports from the Czech Operational Programme Research, Development, and Education, under grant agreement No. CZ.02.1.01/0.0/0.0/15003/0000421.
%
%
%
\bibliographystyle{splncs04}
\bibliography{bibliography}
\end{document}